\documentclass[10pt,twocolumn,letterpaper]{article}
\usepackage[accsupp]{axessibility}
\usepackage{cvpr}
\usepackage{times}
\usepackage{epsfig}
\usepackage{graphicx}
\usepackage{amsmath}
\usepackage{amssymb}

\usepackage{framed}

\usepackage{amssymb}
\usepackage{latexsym}
\usepackage{epsfig}
\usepackage{float}
\usepackage{booktabs,multirow, multicol}

\usepackage{rotating}
\usepackage{epstopdf}
\usepackage{graphicx}
\usepackage{hyperref}

\usepackage{array, longtable}
\usepackage{wrapfig}
\usepackage{enumitem}
\usepackage{tabularx}
\usepackage{url}
\usepackage[altpo]{backnaur}

\usepackage{epsfig}
\usepackage{graphicx}
\usepackage{amsmath}
\usepackage{amssymb}
\usepackage{gensymb}
\usepackage{footmisc}
\usepackage{pifont}
\PassOptionsToPackage{table}{xcolor}
\usepackage{array}
\usepackage{url}

\usepackage{collcell}
\usepackage{xcolor}
\usepackage{colortbl}

\newcommand{\caim}{CAIM{} }
\newcommand{\hfr}{\textit{HFR}{} }

\newcommand{\fullcaim}{Conditional Adaptive Instance Modulation{} }

\definecolor{Gray}{gray}{0.94} 

\begin{document}

\title{Bridging the Gap: Heterogeneous Face Recognition with Conditional Adaptive Instance Modulation}

\author{Anjith George and S\'ebastien Marcel \\
Idiap Research Institute \\
Rue Marconi 19, CH - 1920, Martigny, Switzerland \\
{\tt\small  \{anjith.george, sebastien.marcel\}@idiap.ch  }
}

\maketitle
\thispagestyle{empty}

\begin{abstract}
Heterogeneous Face Recognition (\hfr\!) aims to match face images across different domains, such as thermal and visible spectra, expanding the applicability of Face Recognition (FR) systems to challenging scenarios. However, the domain gap and limited availability of large-scale datasets in the target domain make training robust and invariant \hfr models from scratch difficult. In this work, we treat different modalities as distinct styles and propose a framework to adapt feature maps, bridging the domain gap. We introduce a novel \fullcaim (\caim\!) module that can be integrated into pre-trained FR networks, transforming them into \hfr networks. The \caim block modulates intermediate feature maps, to adapt the style of the target modality effectively bridging the domain gap. Our proposed method allows for end-to-end training with a minimal number of paired samples. We extensively evaluate our approach on multiple challenging benchmarks, demonstrating superior performance compared to state-of-the-art methods. The source code and protocols for reproducing the findings will be made publicly available.
\end{abstract}


\section{Introduction} 

Face recognition (FR) has become a popular method of access control due to its effectiveness and ease of use. Thanks to convolutional neural networks (CNN), most state of the art FR methods achieve excellent performance in ``in the wild'' conditions and even ``human parity'' in face recognition performance \cite{learned2016labeled}. Standard FR systems operate in a homogeneous domain, which means that enrollment and matching are performed using the same modality, typically with face images captured with an RGB camera. However, in some cases, matching in a heterogeneous environment may be advantageous. For example, near-infrared (NIR) cameras, which are common in mobile phones and surveillance cameras, provide superior performance regardless of lighting conditions and are robust against presentation attacks \cite{li2007illumination,george2022comprehensive}. However, training a face recognition system for NIR images will require a large amount of labeled training data which is often not available.

Heterogeneous Face Recognition (\hfr\!) systems aim to enable cross-domain matching, allowing enrolled RGB images to be matched with NIR (or any other modality) images and eliminating the need for separate modalities to be enrolled \cite{klare2012heterogeneous}. This approach is particularly valuable in scenarios where visible images are difficult to acquire. For instance, thermal images can be used for recognition regardless of illumination conditions, enabling face recognition during both day and night, and even from long distances.  \hfr systems can handle and match facial images from a variety of sources and modalities, enhancing their applicability across a broad range of situations and applications.

It can be seen that, \hfr extends the capabilities of Face Recognition (FR) systems in challenging use cases, such as low light or poor visibility, by leveraging the unique properties of different imaging modalities. HFR methods addresses some of the limitations and expands the potential applications of FR systems.

\begin{figure}[t!]
    \centering
    \includegraphics[width=0.99\linewidth]{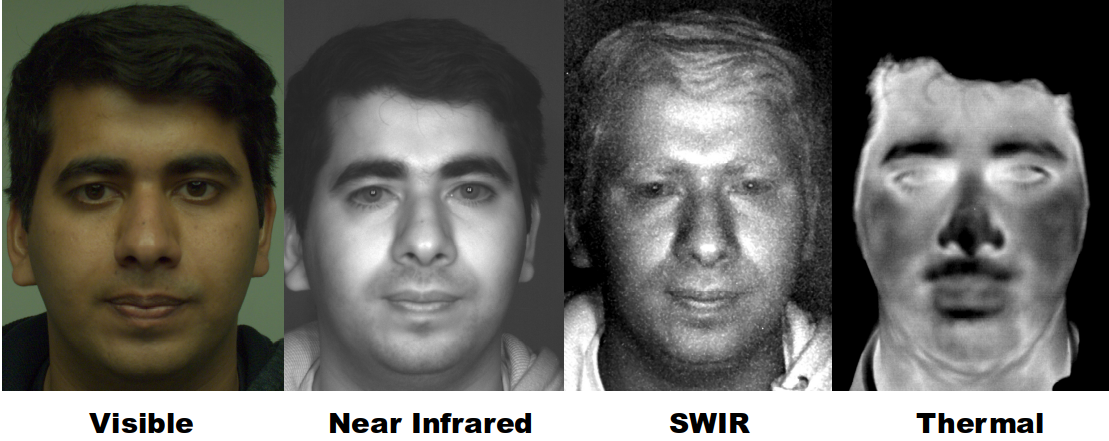}
    \caption{ This figure shows the facial images of the same individual acquired using four distinct imaging modalities (Images taken from  MCXFace dataset \cite{george2022prepended}). The task in \hfr is to facilitate cross-domain matching while overcoming the challenges posed by the domain gap.} \label{fig:hfr}
  \end{figure}

Although highly useful, achieving Heterogeneous Face Recognition (\hfr\!) presents its own challenges. Performing \hfr using images captured in different modalities is difficult due to the domain gap, which often leads to performance degradation when FR networks trained on visible images are combined with other sensing modalities \cite{he2018wasserstein}. Furthermore, training invariant models for both visible and other modalities is challenging due to the limited availability of datasets for new modalities. Collecting large-scale data for these new modalities can also be expensive. Therefore, developing a framework that can be trained with a minimal amount of labeled samples is crucial in this context.

In this work, we utilize a pre-trained face recognition network, trained on a large number of visible spectrum face images, as the base network. We propose to treat different modalities as distinct \textit{styles} and introduce a novel framework to reduce the domain gap. Our method aims to close the gap between visible images and new modalities by modulating the intermediate feature maps of the network. We introduce a new component called \fullcaim (\caim\!) which can be inserted into the intermediate layers of the face recognition network.  This learnable component can be trained end-to-end to convert an FR network to an \hfr network with a minimal amount of training samples.

The main contributions of this work are listed below:

  \begin{itemize}

    \item We formulate the Heterogeneous Face Recognition (\hfr\!) problem as a style modulation problem and address the domain gap in \hfr task using this approach.
          
    \item We introduce a new module called the \fullcaim (\caim\!), which can transform a pre-trained FR network into a heterogeneous face recognition network. The framework can be trained with a minimal number of paired samples from the new modality.
    
    \item We have evaluated the proposed approach in several standard benchmarks to show the effectiveness of our approach.
    
  \end{itemize}
  
  Finally, the protocols and source codes will be made available publicly \footnote{\url{https://gitlab.idiap.ch/bob/bob.paper.ijcb2023_caim_hfr}}.
  
The remainder of the paper is structured as follows: Section \ref{sec:related_work} discusses recent literature on Heterogeneous Face Recognition (\hfr\!). Details of the \caim approach are described in Section \ref{sec:approach}. Extensive evaluation of the \caim approach, along with comparisons with the state-of-the-art, and discussions, are presented in Section \ref{sec:experiments} and \ref{sec:discussions}. 
Lastly, Section \ref{sec:conclusions} presents conclusions and outlines future research directions.

\begin{figure*}[t!]
  \centering
  \includegraphics[width=0.90\linewidth]{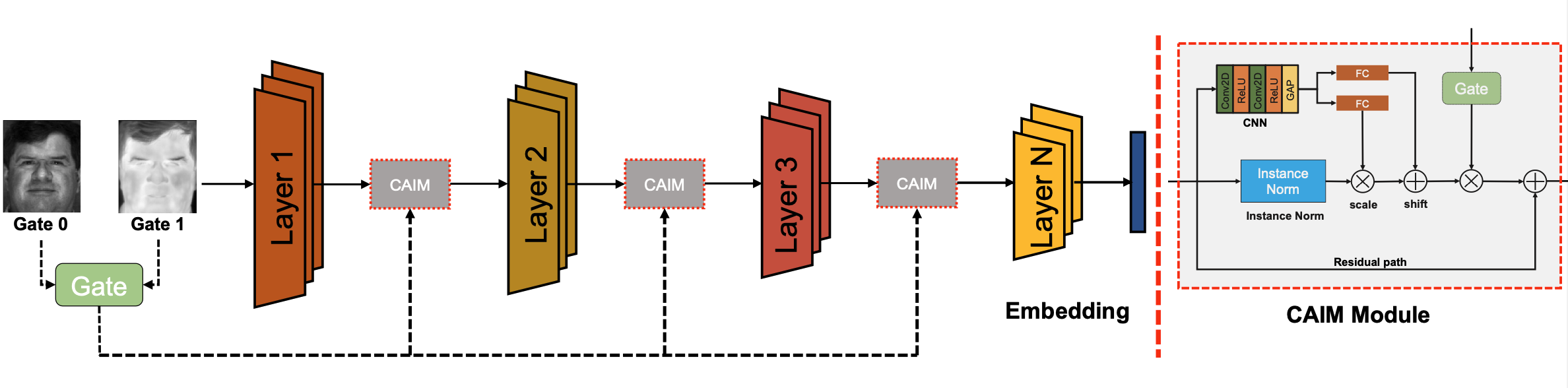}
  \caption{Schematic diagram of the proposed framework: Layer 1 to Layer N represent the frozen blocks of layers from a pretrained Face Recognition (FR) model. Architecture of the \fullcaim (\caim\!) block is shown on the right. The \caim module is inserted between the initial few blocks. }
  \label{fig:framework}
\end{figure*}

\section{Related work} 
\label{sec:related_work}

The goal of \hfr methods is to match images obtained from various sensing modalities. However, due to the domain gap, direct image comparison with a face recognition network degrades performance. Hence, it is essential that \hfr  methods should bridge the gap between modalities. In this section, we review recent literature on strategies proposed for bridging this domain gap.

\subsection{Invariant feature based methods}

Various methods have been proposed for heterogeneous face recognition (\hfr\!) to extract invariant features that can match cross-modal images. Liao \etal \cite{liao2009heterogeneous} proposed a method based on Difference of Gaussian (DoG) filters and multi-scale block local binary patterns (MB-LBP) features. Klare \etal \cite{klare2010matching} proposed a local feature-based discriminant analysis (LFDA) that used scale-invariant feature transform (SIFT) and multi-scale local binary pattern (MLBP) feature descriptors. Zhang \etal \cite{zhang2011coupled} introduced the coupled information-theoretic encoding (CITE) method, which maximizes mutual information between the modalities in the quantized feature spaces. Convolutional neural network (CNN) based methods have also been used for \hfr \cite{he2017learning,he2018wasserstein}. Roy \etal \cite{roy2018novel} proposed a novel method called local maximum quotient (LMQ) to extract invariant characteristics in cross-modal facial images.

\subsection{Common-space projection methods}

The objective of common-space projection methods is to learn a mapping to project different face modalities into a common shared subspace to reduce the domain gap \cite{kan2015multi,he2017learning}. Lin and Tang \cite{lin2006inter} developed a common discriminant feature extraction method for extracting features from cross-modal images and projecting them onto a common feature space. Canonical correlation analysis (CCA) was proposed as a way to match face images between NIR and VIS by Yi \etal \cite{yi2007face}. Regression-based approaches were proposed by Lei \textit{et al.} \cite{lei2009coupled,lei2012coupled} to learn mapping functions that connect cross-modality domains. Sharma and Jacobs \cite{sharma2011bypassing} proposed a partial least squares-based method to learn a linear mapping for face image across different modalities so that the mutual covariance is maximized. Klare and Jain \cite{klare2012heterogeneous} proposed a way to represent face images in terms of their similarity to a set of prototype face images. The prototype-based face representation was then projected onto a linear discriminant subspace, which was used to perform the recognition. In \cite{de2018heterogeneous}, the authors proposed a novel approach to the \hfr task called Domain-Specific Units (DSU). They suggest that high-level features of convolutional neural networks trained on the visible spectrum are domain-independent and can be used to encode images captured in other sensing modalities. The initial layers (DSUs) of a pre-trained FR model are adapted to reduce the domain gap, and the whole pipeline is trained in a contrastive setting. However, the number of layers that need to be adapted is a hyper-parameter that has to be found with an extensive set of experiments.

\subsection{Synthesis based methods}

Synthesis-based methods for \hfr \cite{tang2003face,fu2021dvg} attempt to synthesize the source domain images from the target modality, which enables the use of typical face recognition networks for biometric matching. Wang \etal \cite{wang2008face} proposed a patch-based synthesis approach using Multi-scale Markov Random Fields, while Liu \etal \cite{liu2005nonlinear} used Locally Linear Embedding (LLE) to learn a pixel-level mapping between VIS images and viewed sketches. CycleGAN \cite{zhuUnpairedImagetoImageTranslation2017} has been used to transform images from the target domain to the source domain in \cite{baeNonvisualVisualTranslation2020}. Zhang \etal \cite{zhang2017generative} proposed a Generative Adversarial Network-based Visible Face Synthesis (GAN-VFS) method to synthesize photo-realistic visible face images from polarimetric images. Many recent approaches have been proposed using GANs for the synthesis of VIS images from another modality, such as the Dual Variational Generation (DVG-Face) framework \cite{fu2021dvg}, which achieved state-of-the-art results in many challenging \hfr benchmarks.
In \cite{luo2022memory}, authors proposed a Memory-Modulated Transformer Network (MMTN) for heterogeneous face recognition (\hfr\!) task. They formulated the \hfr problem as a synthesis-based approach, specifically, an unsupervised, reference-based ``one-to-many'' generation task. The MMTN includes a memory module to identify prototypical style patterns and a novel style transformer module for a local fusion of input and reference image styles. Recently, George \etal proposed an approach named Prepended Domain Transformers (PDT) \cite{george2022prepended}, which prepends a neural network module to a pretrained FR network to convert it to an \hfr network. This new block translates the feature representation without explicitly generating the source domain images so that the cross domain embeddings align in the feature space.  

\subsection{Limitations of current approaches}

Recent \hfr methods in the literature \cite{fu2021dvg,zhang2017generative} have predominantly utilized synthesis-based methodologies, especially GAN-based synthesis methods due to their ability to generate high-quality images. The use of a pre-trained FR model in synthesis-based \hfr methods eliminates the need for extensive training data for training a new FR model. However, the synthesis process adds significant computing costs, making it less practical for real-world applications. 

We argue that the domain gap between visible images and other modalities can be viewed as different \textit{styles} and the domain gap can be addressed by modulating the feature maps, obviating the compute and memory overhead for synthesizing source modality images.

\section{Proposed Method}
\label{sec:approach}

We follow the notations consistent with  recent literature \cite{weiss2016survey,de2018heterogeneous, george2022prepended} to formally introduce the \hfr task.

\subsection{Formal definition of \hfr}

Let's consider a domain $\mathcal{D}$ consisting of samples $X \in \mathbb{R}^d$ and a marginal distribution $P(X)$ (of dimensionality $d$). The objective of a face recognition (FR) system, $\mathcal{T}^{fr}$, can be characterized by a label space $Y$ with a conditional probability $P(Y|X,\Theta)$, where $X$ and $Y$ represent random variables, and $\Theta$ denotes the model parameters. During the training phase of an FR system, $P(Y|X, \Theta)$ is generally learned in a supervised manner using a dataset of faces $X={x_1, x_2, ..., x_n}$ along with their associated identities $Y={y_1, y_2, ..., y_n}$.

Now, let's examine a heterogeneous face recognition (\hfr\!) problem. In this scenario, we assume the presence of two domains: a source domain $\mathcal{D}^s = {X^s, P(X^s)}$ and a target domain $\mathcal{D}^t = {X^t, P(X^t)}$, both sharing the labels $Y$.

In general, the goal of the \hfr problem,  $\mathcal{T}^{hfr}$, is to determine a  $\hat{\Theta}$ such that $P(Y|X^s, \Theta) = P(Y|X^t, \hat{\Theta})$. 

\subsection{Proposed Framework}

In our approach, we view face images from different modalities as distinct  \textit{styles} and attribute the domain gap in the \hfr problem as a manifestation of different styles. We hypothesize that, by addressing the domain-specific \textit{style}, we can reduce the domain gap. This is achieved by conditionally modulating the intermediate feature maps of a pre-trained face recognition network. 

Given the parameters of a pre-trained face recognition (FR) model, denoted as $\Theta_{FR}$, originating from the source domain $\mathcal{D}^s$. Instead of adapting the model's weights, we introduce a new set of network blocks, referred to as \caim, which essentially modulate the intermediate feature maps. The \caim module normalizes and adjusts the style of the target modality such that the embeddings of paired samples from both modalities align in the representation space. A schematic diagram of the proposed framework is depicted in Fig. \ref{fig:framework}. The \caim block is inserted into the first few blocks, as they are more closely related to the domain. An external gate activates the \caim module only for the target modality and serves as a pass-through for the reference modality, thus preventing catastrophic forgetting.

Essentially, we can represent the \hfr task in the following way:
\begin{equation}
  P(Y|X_t, \hat{\Theta}) = P(Y|X_t, [\Theta_{FR}, \theta_{\caim_{i,  i \in (1,2,..,K)}}])
\end{equation}

Where, $\theta_{\caim_{i}}$ denotes the $i^{th}$ \caim block out of $K$ blocks.

The only learnable components are $\theta_{\caim_{i}}$, which can be trained in a supervised setting. During training, for source images ($X^s$), the \caim module acts as a pass-through, allowing the reference embedding to be obtained as they pass only through original network layers ($\Theta_{FR}$). For target domain images ($X^t$), they pass through both the original network layers ($\Theta_{FR}$) and the \caim blocks. Contrastive loss \cite{hadsell2006dimensionality} is employed as the loss function during the training phase to ensure the embeddings align in the representation space. The contrastive loss is given as:

\begin{equation}
\begin{split}
\mathcal{L}_{Contrastive}(\hat{\Theta}, Y_p, X_s, X_t)= & (1-Y_p)\frac{1}{2}D_W^{2}  \\
            & + Y_p\frac{1}{2}{max(0, m-D_W)}^{2}
\end{split}
\end{equation}

Where $\hat{\Theta}$ represents the network's weights together with the frozen weights, $X_s$ and $X_t$ denote heterogeneous image pairs. The label $Y_p$ indicates whether the pairs share the same identity. The margin is represented by $m$, and $D_W$ signifies the distance function between the embeddings of the two samples. If the subjects in $X_s$ and $X_t$ have the same identity, the label $Y_p$ is set to $0$, and it is set to $1$ otherwise. The distance function $D_W$ can be calculated using the Euclidean distance (or cosine distance) between the representations extracted by the network. The specifics of the \caim block are elaborated upon in the following subsections.

\subsection{Architecture of the \caim block}

The components of the proposed \fullcaim module are depicted in Figure \ref{fig:framework}. The \caim block is integrated between the frozen layers of a pre-trained facial recognition network. The \caim block accepts an input feature map and a global gate signal, generating an output feature map of identical dimensions. The initial component of the \caim block is an Instance Normalization (IN) block without learnable affine parameters, which serves to normalize each feature map at the sample level. Subsequently, a parallel Convolutional Neural Network (CNN) module is employed to derive a shared representation from the un-normalized feature map. The CNN module comprises two sets of $3\times3$ convolutional layers, each followed by a Rectified Linear Unit (ReLU) activation function, and then a Global Average Pooling (GAP) layer. To obtain new scaling and shift parameters, two fully connected layers are introduced. The fully connected layer estimates the parameters utilized for scaling and shifting the normalized feature maps. A residual connection is then incorporated. When the global gate signal is zero, the entire \caim block functions as an ``Identity'' function, generating un-altered embeddings as the pre-trained FR model for the reference channel.

\subsection{Style Modulation for HFR}

In this section, we expand on  the utilization of \textit{style} modulation as a means to mitigate the domain gap between visible and additional modalities, starting with the Instance Normalization \cite{ulyanov2017improved}.
\subsubsection{Instance normalization}

The statistics of feature maps in deep neural networks (DNNs) have been shown to capture the style of images \cite{gatys2016image}. Ulyanov \etal \cite{ulyanov2017improved} demonstrated that replacing batch normalization layers with Instance Normalization (IN) significantly improves style transfer. The instance normalization layer applies normalization to the feature maps, which can be represented as follows:

\begin{equation}
\mathrm{IN}(x) = \gamma \left(\frac{x - \mu(x)}{\sigma(x)}\right) + \beta
\end{equation}

Here, $\gamma, \beta \in \mathbb{R}^{C}$ represent learned affine parameters from the data, while $\mu(x)$ and $\sigma(x)$ are computed across spatial dimensions for each individual sample (instead of across mini-batches).

In \cite{huang2017arbitrary}, Huang \etal introduced a network module named Adaptive Instance Normalization (AdaIN). This module is designed to align the mean and variance of content features with those of style features in the context of image style transfer. The authors argue that Instance Normalization achieves a form of style normalization by normalizing feature statistics, specifically the mean and variance. 

The AdaIN module takes a content input $x$ and a style input $y$, and aligns the channel-wise mean and variance of $x$ to match those of $y$. AdaIN does not have learnable affine parameters, instead, it adaptively calculates the affine parameters from the style input:

\begin{equation}
\mathrm{AdaIN}(x, y) = \sigma(y) \left(\frac{x - \mu(x)}{\sigma(x)}\right) + \mu(y)
\end{equation}

Here, the normalized content input is scaled by $\sigma(y)$ and shifted by $\mu(y)$. Similar to Instance Normalization, these statistics are computed across spatial locations.

Recent studies \cite{zhou2021domain} have revealed that probabilistically mixing instance-level feature statistics from various source domains can enhance domain generalization. By incorporating diverse styles during training, the model becomes more robust and generalizable across domains. It is important to highlight that, the mixing occurs during the training process of the model incorporating various domains, where the model is trained from scratch. This distinction is crucial, as in the Heterogeneous Face Recognition (HFR) scenario, we begin with a pre-trained face recognition model that has already been trained on the source domain. We extend the idea for \hfr by conditionally modulating the feature maps instead of mixing them.

\subsubsection{\fullcaim}

The AdaIN module is capable of generating or matching images with the style of another image, as discussed above. Specifically, it enables the creation of images that possess the style of another image. In the \hfr scenario, the goal is to modulate the style of target modality images to match that of the visible images to ensure the final embeddings align between modalities. This is especially important given that the pretrained face recognition network has been trained on a large dataset of visible spectrum images.

Consider an intermediate feature map in the face recognition network, denoted by $F \in \mathbb{R}^{CxHxW}$. Here, $C$, $H$, and $W$ represent the number of channels, height, and width of the feature map, respectively.

For the target modality, we would like to modulate this feature maps such that the output embedding from the network aligns for the source and target modalities. 

To accomplish this, we modulate the intermediate feature map using the \caim block.

\begin{equation}
  \hat{\mathbf{F}} = \mathrm{\caim}(\mathbf{F})
\end{equation}

The \caim block's main component is similar to adaptive instance normalization (AdaIN) as it normalizes and modulates the style of target images. However, instead of utilizing an external image to obtain the style, we estimate the modulation factors from the un-normalized input feature maps using a CNN module. Additionally, we combine this step in a residual fashion while injecting the \caim block to a pretrained network.

To elaborate further, we first estimate a shared representation from the input feature map by utilizing a shallow CNN network with global average pooling.

\begin{equation}
  \mathbf{\xi_{f}} = \mathrm{GAP}\big(\mathrm{CNN}(\mathbf{F})\big)
\end{equation}

The $\sigma_{f}$ and $\mu_{f}$ parameters are estimated from this shared representation with two fully connected (FC) layers:
\begin{align}
  \mathbf{\sigma_{f}} &= \mathrm{FC}_{\mathrm{\sigma}}(\mathbf{\xi_{f}}) \\
  \mathbf{\mu_{f}} &= \mathrm{FC}_{\mathrm{\mu}}(\mathbf{\xi_{f}})
\end{align}

The estimated parameters are utilized to scale and shift the normalized feature maps:

\begin{equation}
  \mathrm{AIM}(\mathbf{F}) = \mathbf{\sigma_{f}} \left(\frac{\mathbf{F} - \boldsymbol{\mu}(\mathbf{F})}{\boldsymbol{\sigma}(\mathbf{F})}\right) + \mathbf{\mu_{f}}
\end{equation}

To ensure stable training, we incorporate a residual connection in the proposed framework. Additionally, when incorporating this module, a gate is added to activate the module exclusively for the target modality, leaving the feature maps of the source modality unaltered.

The \caim block can be represented as following:

\begin{equation}
  \mathrm{\caim}(\mathbf{F}, \mathbf{g}) = \mathbf{g} \cdot \mathrm{AIM}(\mathbf{F}) + \mathbf{F}
\end{equation}

Where, $g$ denote the gate, $\mathbf{g} = 1$ for the target modality and $\mathbf{g} = 0$ for the source modality (visible images).

\subsection{Face Recognition backbone}

In the interest of reproducibility, we utilized the publicly accessible pre-trained \textit{Iresnet100} face recognition model provided by Insightface \cite{insightface}. The model was trained on the MS-Celeb-1M dataset \footnote{\url{http://trillionpairs.deepglint.com/data}}, which includes over 70,000 identities. The pre-trained face recognition model accepts three-channel images at a resolution of $112 \times 112$ pixels. Before passing through the FR network, faces are aligned and cropped to ensure eye center coordinates align with predetermined points. In instances of single-channel input data (e.g., NIR, thermal), the single channel is replicated across all three channels without modifying the network architecture. 

\subsection{Implementation details}
The \fullcaim (\caim\!) block is trained using a contrastive learning paradigm, employing contrastive loss within a Siamese network setting \cite{hadsell2006dimensionality}. In all experiments, the margin parameter is set to 2.0. We utilized the Adam Optimizer with a learning rate of $0.0001$ and trained the model for 50 epochs using a batch size of 90. The framework was implemented in PyTorch, leveraging the Bob library \cite{bob2017,bob2012} \footnote{\url{https://www.idiap.ch/software/bob/}}. The frozen layers of the pre-trained face recognition network are shared between both source and target branches. The \caim block is integrated between the frozen layers and is only activated for the target modality when the global gate signal is set to one ($gate=1$). When reference channel images (visible) are processed ($gate=0$), the entire \caim block functions as a pass-through via the residual branch. Only the parameters of the \caim block are updated during training. The experiments are reproducible, and the source code and protocols will be made available publicly.

\section{Experiments}
\label{sec:experiments}

In this section, we present the results of an extensive set of experiments conducted with the \caim framework. The primary focus of these experiments was to evaluate the performance of the \caim approach for VIS-Thermal \hfr across different datasets. Furthermore, we compared the performance of the \caim approach against other heterogeneous settings such as VIS-Sketch and VIS-Low Resolution VIS. In all our experiments, we used the standard cosine distance for comparison.

\subsection{Databases and Protocols}

The following section describes the datasets used in the evaluations.

\textbf{Tufts face dataset:} The Tufts Face Database \cite{panetta2018comprehensive} is a collection of face images from various modalities for the \hfr task. For our evaluation of VIS-Thermal \hfr performance, we use the thermal images provided in the dataset. The dataset comprises 113 identities, consisting of 39 males and 74 females from different demographic regions, and includes images from different modalities for each subject. We adopt the same procedure as in \cite{fu2021dvg}, selecting 50 identities at random for the training set and using the remaining subjects for the test set. We report Rank-1 accuracies and Verification rates at false acceptance rates (FAR) of 1\% and 0.1\% for comparison.

\textbf{MCXFace Dataset:} The MCXFace Dataset \cite{george2022prepended} includes images of 51 individuals captured in various illumination conditions and three distinct sessions using different channels. The channels available include RGB color, thermal, near-infrared (850 nm), short-wave infrared (1300 nm), Depth, Stereo depth, and depth estimated from RGB images. All channels are spatially and temporally registered across all modalities. Five different folds were created for each of the protocols by randomly dividing the subjects into $train$ and $dev$ partitions. Annotations for the left and right eye centers for all images are also included in the dataset. We have performed the evaluations on the challenging ``VIS-Thermal'' protocols of this dataset.

\textbf{Polathermal dataset:} The Polathermal dataset \cite{hu2016polarimetric} is an \hfr dataset collected by the U.S. Army Research Laboratory (ARL). It contains polarimetric LWIR imagery together with color images for 60 subjects. The dataset has conventional thermal images and polarimetric images for each subject. For our experiments, we use the conventional thermal images and follow the five-fold partitions introduced in \cite{de2018heterogeneous}. Specifically, 25 identities are used for training, while the remaining 35 identities are used for testing. We report the average Rank-1 identification rate from the evaluation set of the five folds.

\textbf{SCFace dataset:} The SCFace dataset \cite{grgic2011scface} consists of high-quality enrollment images for face recognition, while the probe samples are low-quality images from various surveillance scenarios captured by different cameras. There are four different protocols in the dataset, based on the quality and distance of the probe samples: close, medium, combined, and far, with the ``far'' protocol being the most challenging. In total, the dataset contains 4,160 static images from 130 subjects (captured in both visible and infrared spectra).

\textbf{CUFSF dataset:} The CUHK Face Sketch FERET Database (CUFSF) \cite{zhang2011coupled} consists of 1194 faces from the FERET dataset \cite{phillips1998feret}, where each face image has a corresponding sketch drawn by an artist. Due to the exaggerations in the sketches, this dataset poses a challenge for the \hfr task. Following \cite{fang2020identity}, we use 250 identities for training the model, and reserve the remaining 944 identities for testing. The Rank-1 accuracies are reported for comparison.

\subsection{Metrics}

We evaluate the models using various performance metrics that are commonly used in previous literature, including Area Under the Curve (AUC), Equal Error Rate (EER), Rank-1 identification rate, and Verification Rate at different false acceptance rates (0.01\%, 0.1\%, 1\%, and 5\%).

\subsection{Experimental results}

The experiments performed in the different datasets and the results are discussed in this section. For comparison, we comparing with state of the art, we compared the results with \caim against the paper baselines reported in \cite{george2022prepended}.

\subsubsection{\textbf{Experiments with Tufts face dataset}}

The performance of the \caim method and other state-of-the-art techniques in the VIS-Thermal protocol of the Tufts face dataset is presented in Table \ref{tab:tufts}. This dataset is very challenging due to variations in pose and other factors. The extreme yaw angles present in the dataset cause a decline in the performance of even visible spectrum face recognition systems, along with a similar decline in \hfr performance. Despite this challenge, the \caim approach achieves the best verification rate and ranks second in Rank-1 accuracy (73.07\%), following DVG-Face \cite{fu2021dvg}. These results demonstrate the effectiveness of the proposed method.

\begin{table}[h]
  \centering
  \caption{Experimental results on VIS-Thermal protocol of the Tufts Face dataset.}
  \label{tab:tufts}
  \resizebox{0.95\columnwidth}{!}{
  \begin{tabular}{lccc}
    \toprule
    \textbf{Method} & \textbf{Rank-1} & \textbf{VR@FAR=1$\%$} & \textbf{VR@FAR=0.1$\%$}  \\ \midrule
      LightCNN \cite{Wu2018ALC} & 29.4 & 23.0 & 5.3 \\
      DVG \cite{fu2019dual} & 56.1 & 44.3 & 17.1 \\
      DVG-Face \cite{fu2021dvg} & \textbf{75.7} & 68.5 & 36.5 \\ 
      DSU-Iresnet100 \cite{george2022prepended} & 49.7 & 49.8 & 28.3 \\   
      
      PDT \cite{george2022prepended} & 65.71 & 69.39 & 45.45 \\ \midrule
      \rowcolor{Gray}
      \textbf{\caim (Proposed)} & 73.07 &\textbf{76.81} & \textbf{46.94} \\
      \bottomrule
  
  \end{tabular}
  }
\end{table}
\vspace{-5mm}

\subsubsection{\textbf{Experiments with MCXFace dataset}}

Table \ref{tab:mcxface} presents the average performance across five folds for the VIS-Thermal protocols in the MCXFace dataset. The reported values are the mean of the five folds in the dataset. The baseline model shown corresponds to the performance of the pretrained \textit{Iresnet100} FR model directly on the thermal images.  It can be seen that the proposed \caim approach achieves the best performance compared to other methods with an average Rank-1 accuracy of 87.24 \%.

\begin{table}[h]
  \caption{Performance of the proposed approach in the VIS-Thermal protocol of  MCXFace dataset, the Baseline is a pre-trained \textit{Iresnet100} model. }
  \label{tab:mcxface}
  \centering
  \resizebox{0.98\columnwidth}{!}{%
  \begin{tabular}{lrrr}
  \toprule
  \textbf{Method} & \textbf{AUC}   & \textbf{EER}   & \textbf{Rank-1}   \\ \midrule
 Baseline & 84.45 $\pm$ 3.70  & 22.07 $\pm$ 2.81 & 47.23 $\pm$ 3.93    \\
DSU-Iresnet100 \cite{george2022prepended} & 98.12 $\pm$ 0.75 & 6.58 $\pm$ 1.35 & 83.43 $\pm$ 5.47 \\

PDT \cite{george2022prepended}      & 98.43 $\pm$ 0.78  &  6.52 $\pm$ 1.45  & 84.52 $\pm$ 5.36   \\ \midrule

\rowcolor{Gray}
\textbf{\caim (Proposed)}  & \textbf{98.97 $\pm$ 0.24} & \textbf{5.05 $\pm$ 0.91} & \textbf{87.24$\pm$2.75} \\
\bottomrule
  \end{tabular}
  }
  \end{table}
\vspace{-5mm}

\subsubsection{\textbf{Experiments with Polathermal dataset}}
We have performed experiments in the thermal to visible recognition scenarios in the Polathermal dataset and the results are presented in Table \ref{tab:polathermal}. The table shows the average Rank-1 identification rate in the five protocols of the Polathermal `thermal to visible protocols' (using the reproducible protocols in \cite{de2018heterogeneous}). The proposed \caim approach achieves an average Rank-1 accuracy of 95.00\% with a standard deviation of (1.63\%), only second to PDT approach \cite{george2022prepended}. 

\begin{table}[ht]
\caption{Pola Thermal - Average Rank-1 recognition rate}
\label{tab:polathermal}
\begin{center}
  \resizebox{0.7\columnwidth}{!}{
  \begin{tabular}{lr}
    \toprule
    \textbf{Method} & \textbf{Mean (Std. Dev.)} \\ \midrule
    
    DPM in \cite{hu2016polarimetric}   & 75.31 \% (-) \\ 
    CpNN in \cite{hu2016polarimetric}  & 78.72 \% (-) \\ 
    PLS in \cite{hu2016polarimetric}   & 53.05\% (-)  \\  \midrule

    LBPs + DoG in \cite{liao2009heterogeneous} & 36.8\% (3.5) \\ 
    ISV in \cite{de2016heterogeneous}       & 23.5\% (1.1) \\ 
    GFK in \cite{sequeira2017cross}             & 34.1\% (2.9) \\ 

    DSU(Best Result) \cite{de2018heterogeneous} & 76.3\% (2.1) \\

    DSU-Iresnet100 \cite{george2022prepended} & 88.2\% (5.8) \\
    PDT \cite{george2022prepended} & \textbf{97.1\% (1.3)} \\ \midrule
    \rowcolor{Gray}
    \textbf{\caim (Proposed)} & 95.00\% (1.63) \\
    \bottomrule 
  \end{tabular}
  }
\end{center}
\end{table}
\vspace{-5mm}

\subsubsection{\textbf{Experiments with SCFace dataset}}

We conducted a series of experiments on the SCFace dataset to evaluate the performance of the proposed approach using the visible images protocol. The dataset presents a heterogeneity challenge due to the quality disparity between the gallery (high-resolution mugshots) and probe (low-resolution surveillance camera) images. The results are presented in Table \ref{tab:scface} and are based on the evaluation set of the standard protocols. The baseline model employed in this experiment is a pre-trained \textit{Iresnet100} model, while the proposed \caim model is trained using contrastive training. It can be seen that the performance of the baseline model improves with the proposed approach in most of the cases. In particular, the improvement is more significant in the ``far'' protocol where the quality of the probe images is very low. The \caim module helps in adapting the intermediate feature map so that the \hfr framework is invariant to quality and resolution, leading to improved results compared to the baseline. The proposed method achieves comparable performance to the PDT approach in this dataset.

\begin{table}[h]
  \caption{Performance of the proposed approach in the SCFace dataset, the Baseline is a pretrained \textit{Iresnet100} model. }
  \label{tab:scface}
  \centering
  \resizebox{0.98\columnwidth}{!}{%
  \begin{tabular}{lcrrrr}
  \toprule
  \textbf{Protocol}             & \textbf{Method} & \textbf{AUC}   & \textbf{EER}   & \textbf{Rank-1}    & \begin{tabular}[c]{@{}c@{}} \textbf{VR@}\\\textbf{FAR=0.1\%} \end{tabular} \\ \midrule
  \multirow{3}{*}{Close} & Baseline & 100.0   & 0.00    & \textbf{100.0}   & 100.0                \\
                        &DSU-Iresnet100 \cite{george2022prepended} & 100.0 & 0.00 & \textbf{100.0} & 100.0    \\

                             &  PDT \cite{george2022prepended}  & 100.0   &  0.00    &  \textbf{100.0}   & 100.0                    \\ 
                             & \cellcolor{Gray} \textbf{\caim (Proposed)} &  \cellcolor{Gray} 100.0 &  \cellcolor{Gray} 0.01 &   \cellcolor{Gray} \textbf{100.0} &  \cellcolor{Gray} 100.0 \\ \midrule

\multirow{3}{*}{Medium}   & Baseline & 99.81 & 2.33 & 98.60  & 92.09              \\
&DSU-Iresnet100 \cite{george2022prepended} & 99.95 & 1.39 & 98.98 & 93.25 \\

                             & PDT \cite{george2022prepended} & 99.96 &  0.93 &  \textbf{99.07} & 95.81                 \\ 
                             & \cellcolor{Gray} \textbf{\caim (Proposed)} &  \cellcolor{Gray} 99.92 &  \cellcolor{Gray} 1.86 &   \cellcolor{Gray} 98.60 &  \cellcolor{Gray} 94.88 \\ \midrule
\multirow{3}{*}{Combined}    & Baseline & 98.59 & 6.67 & 91.01 & 77.67          \\
                                &DSU-Iresnet100 \cite{george2022prepended} & 98.91 & 4.96 &92.71 & 80.93  \\
                             & PDT \cite{george2022prepended} & 99.06 & 4.50  & 93.18 & 82.02           \\  
                             & \cellcolor{Gray} \textbf{\caim (Proposed)} &  \cellcolor{Gray} 99.58 &  \cellcolor{Gray} 3.24 &  \cellcolor{Gray} \textbf{94.57} &  \cellcolor{Gray} 84.65 \\    \midrule
\multirow{3}{*}{Far}     & Baseline & 96.59 & 9.37 & 74.42 & 49.77           \\
                                &DSU-Iresnet100 \cite{george2022prepended} & 97.18 & 8.37 & 79.53 & 58.26  \\

                             & PDT \cite{george2022prepended}   &  98.31 & 6.98 &  84.19 & 60.00            \\ 
                             & \cellcolor{Gray} \textbf{\caim (Proposed)} &  \cellcolor{Gray} 98.81 &  \cellcolor{Gray} 5.09 &  \cellcolor{Gray} \textbf{86.05} &  \cellcolor{Gray} 61.86 \\  

  \bottomrule
  \end{tabular}
  }
  \end{table}
\vspace{-5mm}

\subsubsection{\textbf{Experiments with CUFSF dataset}}

In this section, we present experiments on the challenging task of sketch to photo recognition. We report the Rank-1 accuracies obtained with the baseline and other methods in Table \ref{tab:cufsf} using the protocols outlined in \cite{fang2020identity}. The proposed approach achieves a Rank-1 accuracy of 76.38\%, which is the best among the compared methods. However, the absolute accuracy in sketch to photo recognition is low compared to other modalities. The CUFSF dataset contains viewed hand-drawn sketch images \cite{klum2014facesketchid} that appear holistically similar to the original subjects for humans. Unlike other imaging modalities such as thermal, near-infrared, and SWIR, sketch images may not preserve the discriminative information that a face recognition network seeks, as they contain exaggerations depending on the artist, making them more challenging for \hfr.  Nevertheless the proposed \caim approach improves the performance significantly.

\begin{table}[ht]
\caption{CUFSF: Rank-1 recognition in sketch to photo recognition}
\label{tab:cufsf}

\begin{center}
\resizebox{0.5\columnwidth}{!}{%
  \begin{tabular}{lrr}
    \toprule
    \textbf{Method} & \textbf{Rank-1} \\ \midrule
    Baseline & 56.57 \\
    IACycleGAN \cite{fang2020identity} &64.94 \\
    DSU-Iresnet100 \cite{george2022prepended} & 67.06 \\ 
    PDT \cite{george2022prepended}  & 71.08 \\ \midrule
    \rowcolor{Gray} 
    \textbf{\caim (Proposed)} & \textbf{76.38} \\
    \bottomrule 
  \end{tabular}

  }
\end{center}
\end{table}

\vspace{-5mm}

\begin{table}[!htb]
  \caption{Performance with different number of \caim blocks. 1-5 indicates the \caim module is inserted in all blocks from first to fifth layers. Experiment performed in Tufts face dataset. }
  \centering
  \resizebox{0.99\columnwidth}{!}{%
\begin{tabular}{lrrrrr}
  \toprule
  \textbf{Layers} &             \textbf{AUC} &             \textbf{EER} &              \textbf{Rank-1} &          \textbf{VR(0.1\% FAR)} &            \textbf{VR(1\% FAR)}    \\
  \midrule
  1 &  91.28 &  17.10 &  49.19 &    3.34 &  49.17  \\
  1-2 &  94.91 &  11.35 &  64.45 &   39.70 &  67.72  \\
  1-3 &  \textbf{97.01} &   \textbf{8.53} &  \textbf{73.07} &   \textbf{46.94} &  \textbf{76.81} \\
  1-4 &  96.18 &   9.28 &  68.76 &   45.08 &  72.36 \\
  1-5 &  95.73 &  10.76 &  69.30 &   33.40 &  71.61 \\
  \bottomrule
  \end{tabular}
  }
  \label{tab:ablation_layer}
\end{table}

\vspace{-5mm}

\subsection{Ablation Study}
To understand the effect of having a different number of \caim blocks, we performed a set of experiments in the Tufts face dataset by inserting a different number of \caim blocks in the pre-trained FR network.  We start by placing only one \caim block after the first block of the pre-trained FR layer. Then we increased the number of \caim blocks from one to 5.  
The results of this experiment are presented in Table. \ref{tab:ablation_layer}. Our analysis reveals that adapting feature maps in the lower layers of the network is more beneficial, as they tend to be more related to the modality.  In this case,  adding three \caim blocks achieved the best performance (this setting is used in all other experiments). Conversely, adapting more layers does not bring significant improvements as they are more task-specific. In our case, the task is face recognition which is the same for both source and target modalities.

Further to understand the effectiveness of the conditional operation, we conducted experiments using the \textbf{AIM} and Instance Norm (IN) modules in an unconditional manner. These experiments were conducted using the Tufts-face dataset, with the results shown in Table \ref{tab:tufts_uncond}. The conditional path in \caim  keeps the original performance on the source modality intact and prevent catastrophic forgetting when adapted to an extra modality. It can be seen that an unconditional integration of the block violates this premise and leads to inferior performance. These results underline both the effectiveness and necessity of the conditional operation.

\begin{table}[h]
  \centering
  \caption{Ablation experiments on Tufts Face dataset with unconditional block.}
  \label{tab:tufts_uncond}
  \resizebox{0.95\columnwidth}{!}{
  \begin{tabular}{lccc}
    \toprule
    \textbf{Method} & \textbf{Rank-1} & \textbf{VR@FAR=1$\%$} & \textbf{VR@FAR=0.1$\%$}  \\ \midrule

AIM & 6.82  &  3.71 &   0.19 \\

IN &36.27  &  17.44 &   3.53 \\ \midrule
\rowcolor{Gray}
      \textbf{\caim (Proposed)} & 73.07 &\textbf{76.81} & \textbf{46.94} \\
      \bottomrule
  
  \end{tabular}
  }
\end{table}

\section{Discussions}
\label{sec:discussions}
We propose a novel approach that modulates the target modality feature maps to align with the style of visible images, effectively bridging the gap between different modalities. To achieve this, we introduce a novel module called \caim that can be inserted into a pre-trained FR model, which enables the conversion of a face recognition model to an \hfr model. Our experimental results demonstrate the effectiveness and robustness of our proposed approach, with state-of-the-art performance achieved in various \hfr benchmarks. In four out of five datasets, the proposed approach outperforms all other approaches compared. Our method shows superior adaptability in the feature space compared to PDT \cite{george2022prepended}, whose transformations are constrained by the PDT block’s receptive field, making our framework more flexible. The proposed approach can be extended to newer FR architectures, and can also be improved by better training methods.
\section{Conclusions}
\label{sec:conclusions}

In this work, we present a novel framework for heterogeneous face recognition by treating different domains as different \textit{styles}. Our proposed approach can convert a face recognition (FR) model to an \hfr model by modulating the style of target modality feature maps to match that of visible images. To achieve this, we introduce a novel network module called \caim\!, which can be inserted between the frozen layers of a pre-trained FR network. This combined module can be trained for \hfr using contrastive training. Our experimental results demonstrate the state-of-the-art performance of our proposed method in various challenging benchmarks, indicating its effectiveness and robustness. The source codes and protocols will be made publicly available to facilitate the extension of our work.

\section*{Acknowledgment}

The authors would like to thank the Swiss Center for Biometrics Research and Testing for supporting the research leading to results published in this paper.

\clearpage
{\small
\bibliographystyle{ieee}
\bibliography{sn-bibliography}
}

\end{document}